\documentclass[10pt,twocolumn,letterpaper]{article}

\usepackage{iccv}
\usepackage{times}
\usepackage{epsfig}
\usepackage{graphicx}
\usepackage{amsmath}
\usepackage{amssymb}

\usepackage[algo2e]{algorithm2e}
\usepackage{setspace}
\usepackage{subfigure}
\usepackage{booktabs}
\usepackage{graphics}
\usepackage{algorithm}
\usepackage{algorithmic}
\usepackage{mathrsfs}
\usepackage{booktabs}
\usepackage{bm}
\usepackage{bbm}
\usepackage{float}
\usepackage{url}
\usepackage{amsfonts}
\usepackage[square,sort&compress,numbers]{natbib}
\usepackage{multirow}
\usepackage{textcomp}
\usepackage{supertabular}
\usepackage{array}
\newcommand*\samethanks[1][\value{footnote}]{\footnotemark[#1]}
\newcolumntype{L}[1]{>{\raggedright\arraybackslash}p{#1}}
\newcolumntype{C}[1]{>{\centering\arraybackslash}p{#1}}
\newcolumntype{R}[1]{>{\raggedleft\arraybackslash}p{#1}}

\usepackage[breaklinks=true,bookmarks=false]{hyperref}

\iccvfinalcopy 

\def\iccvPaperID{2957} 

\ificcvfinal\fi

\begin{document}

\title{Face-to-Parameter Translation for Game Character Auto-Creation}

\author{Tianyang Shi\\
NetEase Fuxi AI Lab\\
{\tt\small shitianyang@corp.netease.com}
\and
Yi Yuan\thanks{Corresponding authors}\\
NetEase Fuxi AI Lab\\
{\tt\small yuanyi@corp.netease.com}
\and
Changjie Fan\\
NetEase Fuxi AI Lab\\
{\tt\small fanchangjie@corp.netease.com}
\and
Zhengxia Zou\\
University of Michigan, Ann Arbor\\
{\tt\small zzhengxi@umich.edu}
\and
Zhenwei Shi\samethanks\\
Beihang University\\
{\tt\small shizhenwei@buaa.edu.cn}
\and
Yong Liu\\
Zhejiang University\\
{\tt\small yongliu@iipc.zju.edu.cn}
}

\maketitle
\ificcvfinal\thispagestyle{empty}\fi

\begin{abstract}
Character customization system is an important component in Role-Playing Games (RPGs), where players are allowed to edit the facial appearance of their in-game characters with their own preferences rather than using default templates. This paper proposes a method for automatically creating in-game characters of players according to an input face photo. We formulate the above ``artistic creation'' process under a facial similarity measurement and parameter searching paradigm by solving an optimization problem over a large set of physically meaningful facial parameters. To effectively minimize the distance between the created face and the real one, two loss functions, i.e.\ a ``discriminative loss'' and a ``facial content loss'', are specifically designed. As the rendering process of a game engine is not differentiable, a generative network is further introduced as an ``imitator'' to imitate the physical behavior of the game engine so that the proposed method can be implemented under a neural style transfer framework and the parameters can be optimized by gradient descent. Experimental results demonstrate that our method achieves a high degree of generation similarity between the input face photo and the created in-game character in terms of both global appearance and local details. Our method has been deployed in a new game last year and has now been used by players over 1 million times.\par
\end{abstract}

\section{Introduction}

\begin{figure}
  \centering
  \includegraphics[width=0.9\linewidth]{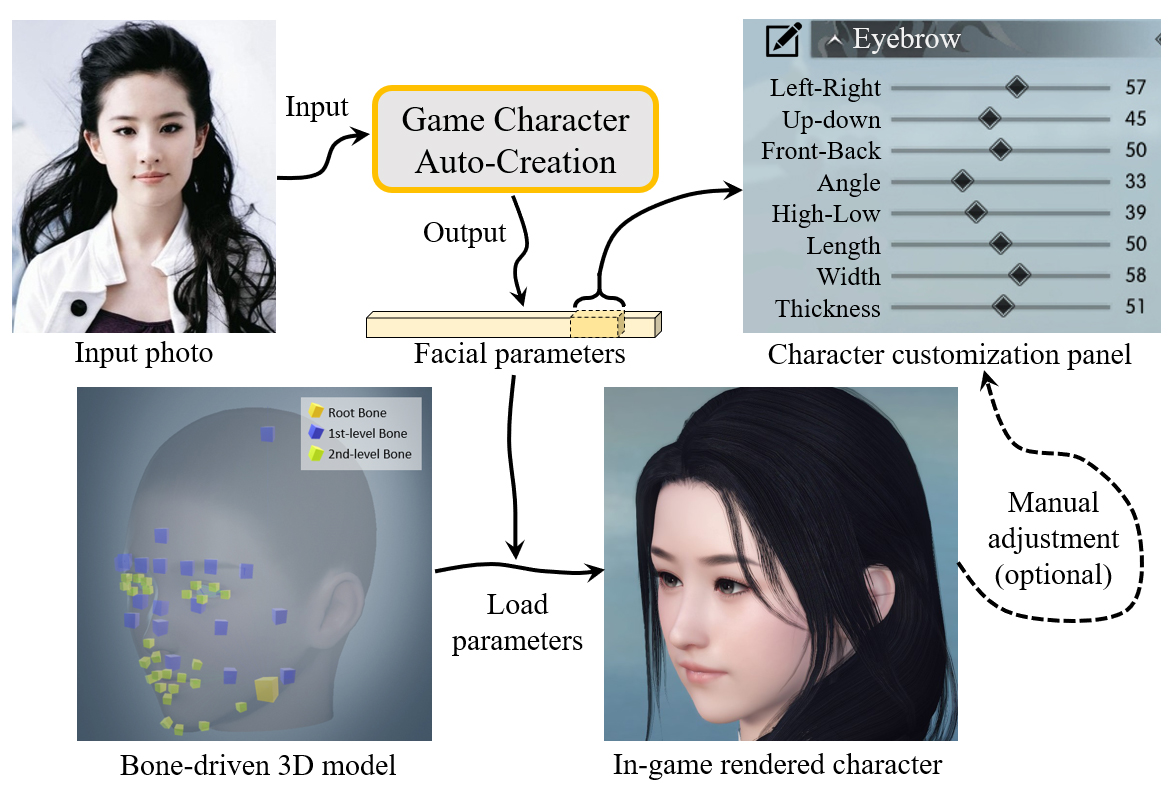}
  \caption{An overview of our method: We propose a method for game character auto-creation based on an input face photo, which can be formulated under a facial similarity measurement and the searching of a large set of physically meaningful facial parameters. Users can further optionally fine-tune the facial parameters on our creation according to their needs.}
  \label{overview}
\end{figure}

The character customization system is an important component in role-playing games (RPGs), where players are allowed to edit the profiles of their in-game characters according to their own preferences (e.g.\ a pop star or even themselves) rather than using default templates. In recent RPGs, to improve the player's immersion, character customization systems are becoming more and more sophisticated. As a result, the character customization process turns out to be time-consuming and laborious for most players. For example, in ``Grand Theft Auto Online\footnote{https://www.rockstargames.com/GTAOnline}'', ``Dark Souls III\footnote{https://www.darksouls.jp}'', and ``Justice\footnote{https://n.163.com}'', to create an in-game character with a desired facial appearance according to a real face photo, a player has to spend several hours manually adjusting hundreds of parameters, even after considerable practice. \par 

A standard work-flow for the creation of the character's face in RPGs begins with the configuration of a large set of facial parameters. A game engine then takes in these user-specified parameters as inputs and generates the 3D faces. Arguably, the game character customization can be considered as a special case of a ``monocular 3D face reconstruction'' \cite{blanz1999morphable, Richardson_2017_CVPR, Tewari_2017_ICCV} or a ``style transfer'' \cite{Efros2001Image, Elad2017Style, Gatys2016Image} problem. Generating the semantic content and 3D structures of an image have long been difficult tasks in the computer vision field. In recent years, thanks to the development of deep learning techniques, computers are now able to automatically produce images with new styles \cite{Gatys2016Image, Li2016Local, Gu_2018_CVPR} and even generate 3D structures from single facial image \cite{Richardson_2017_CVPR, Tewari_2017_ICCV,Tran_2017_CVPR} by taking advantages of deep Convolutional Neural Networks (CNN). \par

But unfortunately, the above methods cannot be directly applied in the game environment. The reasons are threefold. First, these methods are not designed to generate parameterized characters, which is essential for most game engines as they usually take in the customized parameters of a game character rather than images or 3D meshgrids. Second, these methods are not friendly to user interactions as it is extremely difficult for most users to directly edit the 3D meshgrids or rasterized images. Finally, the rendering process of a game engine given a set of user-specified parameters is \textit{not differentiable}, which has further restricted the applicability of deep learning methods in the game environment. \par

Considering the above problems, this paper proposes a method for the automatic creation of an in-game character according to a player's input face photo, as shown in Fig.\ \ref{overview}. We formulate the above ``artistic creation'' process under a facial similarity measurement and a parameter searching paradigm by solving an optimization problem over a large set of facial parameters. Different from previous 3D face reconstruction approaches \cite{blanz1999morphable, Richardson_2017_CVPR, Tewari_2017_ICCV} that produce 3D face meshgrids, our method creates 3D profile for a bone-driven model by predicting a set of facial parameters with a clear physical significance. In our method, each of the parameters controls an individual attribute of each facial components, including the position, orientation and scale. More importantly, our method supports additional user interactions on the basis of the creating results, where players are allowed to make further improvements on their profiles according to their needs. As the rendering process of a game engine is not differentiable, a generative network $G$ is designed as an ``imitator'' to imitate the physical behavior of the game engine, so that the proposed method can be implemented under a neural style transfer framework and the facial parameters can be optimized by using gradient descent, thus we refer to our method as a ``Face-to-Parameter (F2P)'' translation method.\par

As the facial parameter searching in our method is essentially a cross-domain image similarity measurement problem, we take advantages of the deep CNN and multi-task learning to specifically design two kinds of loss functions, i.e.\ a ``discriminative loss'' and a ``facial content loss'' -- the former corresponding to the similarity measurement of the global facial appearance and the latter focusing more on local details. Due to the all CNN design, our model can be optimized in a unified end-to-end framework. In this way, the input photo can be effectively converted to a realistic in-game character by minimizing the distance between the created face and the real one. Our method has been deployed in a new game since Oct.\ 2018 and now has been providing over 1 million times services.\par

Our contributions are summarized as follows:\par

1) We propose an end-to-end approach for face-to-parameter translation and game character auto-creation. To our best knowledge, there are few previous works have been done on this topic.\par 

2) As the rendering process of a game engine is not differentiable, we introduce an imitator by constructing a deep generative network to imitate the behavior of a game engine. In this way, the gradient can smoothly back-propagate to the input so that the facial parameters can be updated by gradient descent.\par

3) Two loss functions are specifically designed for the cross-domain facial similarity measurement. The proposed objective can be jointly optimized in a multi-task learning framework.\par

\begin{figure*}
  \centering
  \includegraphics[width=0.9\linewidth]{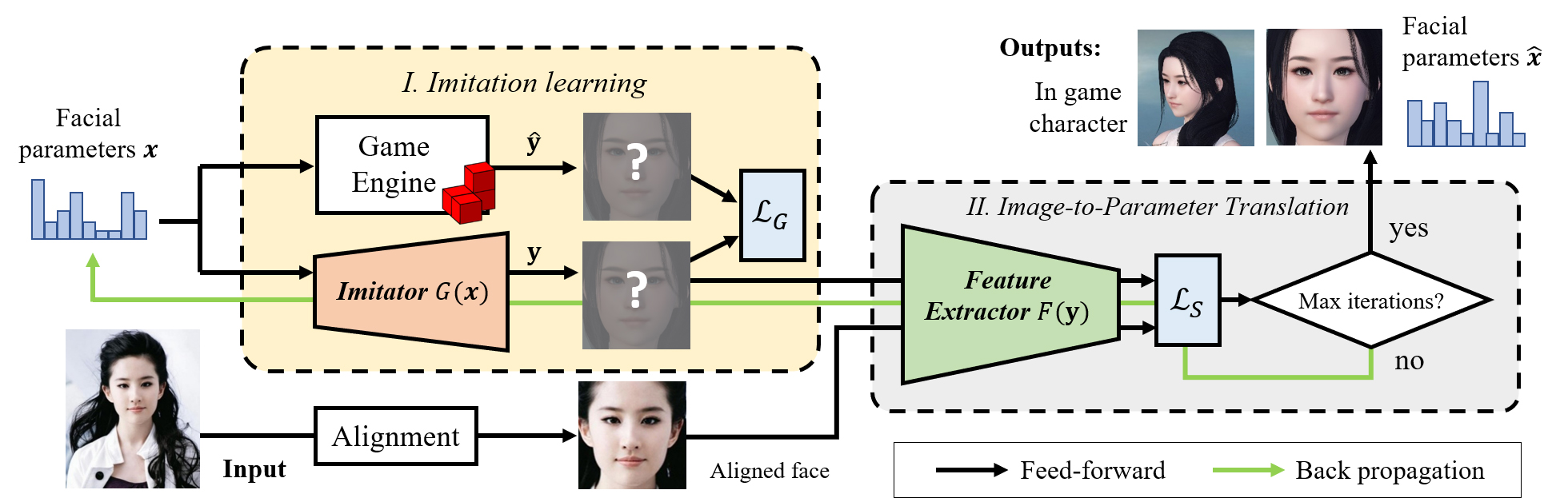}
  \caption{The processing pipeline of the proposed method. Our model consists of an Imitator $G(\bm{x})$ and and a Feature Extractor $F(\bm{y})$. The former aims to simulate the behavior of a game engine by taking in the user-customized facial parameters $\bm{x}$ and producing a ``rendered'' facial image $\bm{y}$. The latter determines the feature space in which the facial similarity measurement can be performed to search for an optimal set of facial parameters.}
  \label{framework}
\end{figure*}

\section{Related work}

{\bf Neural style transfer:} The style transfer from one image onto another has long been a challenging task in image processing  \cite{Efros2001Image, Elad2017Style}. In recent years, Neural Style Transfer (NST) has made huge breakthroughs in style transfer tasks \cite{Efros2001Image, Elad2017Style, Gatys2016Image}, where the integration of deep convolutional features makes it possible to explicitly separate the ``content'' and ``style'' from input images. Most of the recent NST models are designed to minimize the following objective:
\begin{equation}
    \mathcal{L}_{total}  =  \mathcal{L}_{content} + \lambda \mathcal{L}_{style},
\end{equation}
where $\mathcal{L}_{content}$ and $\mathcal{L}_{style}$ correspond to the constraints on the image content and the image style. $\lambda$ controls the balance of the above two objectives. \par

Current NST methods can be divided into two groups: the global methods \cite{Gatys2016Image, Lu2017Decoder, johnson2016perceptual, li2017universal, Sendik2017Correlations, dumoulin2017learned, huang2017arbitrary, risser2017stable, Chen_2017_CVPR} and the local methods \cite{Li2016Local, chen2016fast, Liao2017Attribute}, where the former measures the style similarity based on the global feature statistics, while the latter performs the patch-level matching to better preserve the local details. To integrate both advantages of global and local methods, the hybrid method \cite{Gu_2018_CVPR} is proposed more recently. However, these methods are specifically designed for image-to-image translation rather than a bone-driven 3D face model thus cannot be applied to in-game environments.\par

{\bf Monocular 3D face reconstruction:} Monocular 3D face reconstruction aims to recover the 3D structure of the human face from a single 2D facial image. The traditional approaches of this group are the 3D morphable model (3DMM) \cite{blanz1999morphable} and its variants \cite{Peng_2017_CVPR, Booth_2017_CVPR}, where a 3D face model is first parameterized \cite{BFM2017} and is then optimized to match a 2D facial image. In recent years, deep learning based face reconstruction methods are now able to achieve end-to-end reconstruction from a 2D image to 3D meshgrids \cite{Tran_2017_CVPR,Richardson_2017_CVPR, Tewari_2017_ICCV, Dou_2017_CVPR, Extreme_2018_CVPR, Nonlinear_2018_CVPR}. However, these methods are not friendly for user interactions since it is not easy to edit on 3d meshgrids, and their generated face parameters lack explicit physical meanings. A similar work to ours is Genova's ``differentiable renderer'' \cite{genova2018unsupervised}, in which they directly render parameterized 3D face model by employing a differentiable rasterizer. In this paper, we introduce a more unified solution to differentiate and imitate a game engine regardless of the type of its renderer and 3D model structure by using a CNN model.
\par

{\bf Generative Adversarial Network (GAN):} In addition to the above approaches, GAN \cite{NIPS2014_5423} has made great progress in image generation \cite{mirza2014conditional, radford2015unsupervised, arjovsky2017wasserstein, Lu_2018_ECCV}, and has shown great potential in image style transfer tasks \cite{NIPS2016_6544, Yi_2017_ICCV, Zhu_2017_ICCV, Isola_2017_CVPR, Chang_2018_CVPR}. A similar approach with our method is Tied Output Synthesis (TOS) \cite{wolf2017unsupervised}, which takes advantage of adversarial training to create parameterized avatars based on human photos. However, this method is designed to predict discrete attributes rather than continuous facial parameters. In addition, in RPGs, learning to directly predict a large set of 3D facial parameters from a 2D photo will lead to the defect of parameter ambiguity since the intrinsic correspondence from 2D to 3D is an ``one-to-many'' mapping. For this reason, instead of directly learning to predict continuous facial parameters, we frame the face generation under an NST framework by optimizing the input facial parameters to maximize the similarity between the created face and the real one.\par

\section{Method}

Our model consists of an Imitator $G(\bm{x})$ and a Feature Extractor $F(\bm{y})$, where the former aims to imitate the behavior of a game engine by taking in the user-customized facial parameters $\bm{x}$ and producing a ``rendered'' facial image $\bm{y}$, while the latter determines the feature space in which the facial similarity measurement can be performed to optimize the facial parameters. The processing pipeline of our method is shown in Fig.\ \ref{framework}.\par

\subsection{Imitator}

\begin{figure}
  \centering
  \includegraphics[width=0.9\linewidth]{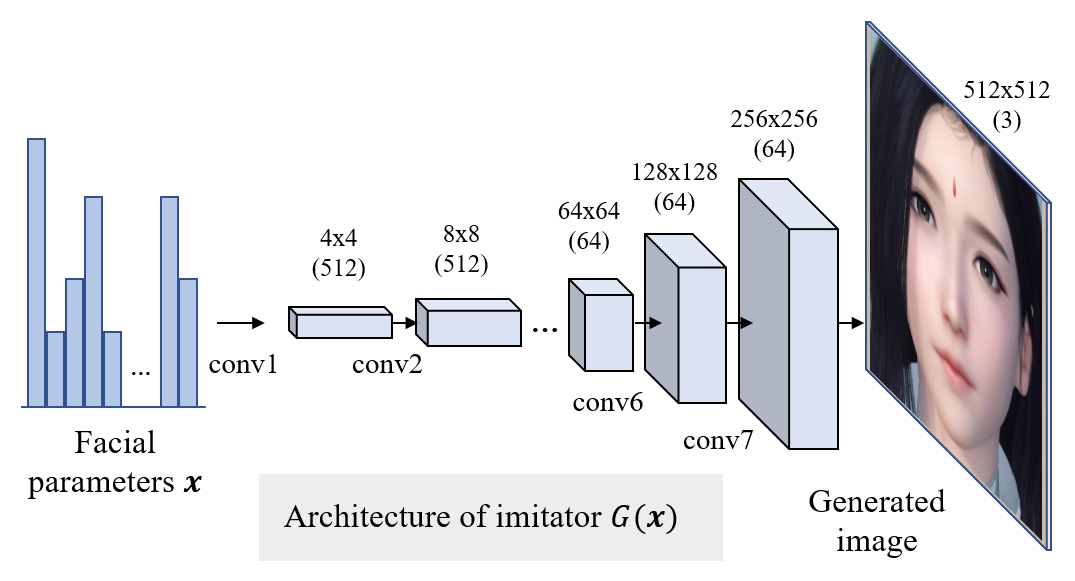}
  \caption{The architecture of our imitator $G(\bm{x})$. We train the imitator to learn a mapping from the input facial customization parameters $\bm{x}$ to the rendered facial image $\hat{\bm{y}}$ produced by the game engine.
  } 
  \label{generator}
\end{figure}

We train a convolutional neural network as our imitator to fit the input-output relationship of a game engine so that to make the character customization system differentiable. We take the similar network configuration of DCGAN \cite{radford2015unsupervised} in our imitator $G(\bm{x})$, which consists of eight transposed convolution layers. The architecture of our imitator is shown in Fig.\ \ref{generator}. For simplicity, our imitator $G$ only fits the front view of the facial model with the corresponding facial customization parameters.\par

We frame the learning and prediction of the imitator as a standard deep learning based regression problem, where we aim to minimize the difference between the in-game rendered image and the generated one in their raw pixel space. The loss function for training the imitator is designed as follows:
\begin{equation}
\begin{split}
    \mathcal{L}_G (\bm{x}) & = E_{\bm{x}\sim u(\bm{x})}\{ {\|\bm{y} - \hat{\bm{y}}\|}_1 \} \\
    & = E_{\bm{x}\sim u(\bm{x})}\{ {\|G(\bm{x}) - \text{Engine}(\bm{x})\|}_1 \},
\end{split}
\end{equation}
where $\bm{x}$ represents the input facial parameters, $G(\bm{x})$ represents the output of the imitator, $\hat{\bm{y}}=\text{Engine}(\bm{x})$ represents the rendering output of the game engine. We use $l_1$ loss function rather than $l_2$ as $l_1$ encourages less blurring. The input parameters $\bm{x}$ are  sampled from a multidimensional uniform distribution $u(\bm{x})$. Finally, we aim to solve: 
\begin{equation}
   G^\star =  \arg \min_{G} \mathcal{L}_G (\bm{x}).   
\end{equation}\par

\begin{figure}
  \centering
  \includegraphics[width=0.9\linewidth]{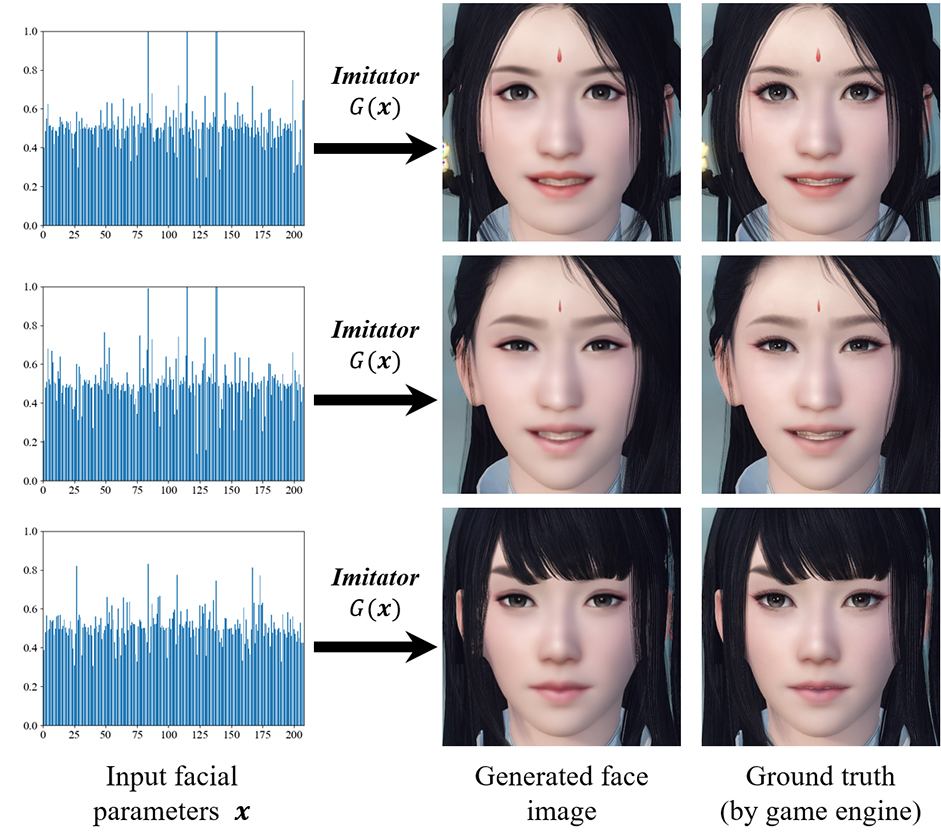}
  \caption{Some examples of the generated face images by our imitator $G(\bm{x})$ and the corresponding ground truth. The facial parameters of these images are created manually.
  }
  \label{exp:generator}
\end{figure}

In the training process, we randomly generate 20,000 individual faces with their corresponding facial customization parameters by using the engine of the game ``Justice''. 80\% face samples are used for training and the rests are used for validation. Fig.\ \ref{exp:generator} shows three examples of the ``rendering'' results of our imitator. The facial parameters of these images are created manually. As the training samples are generated randomly from a unified distribution of facial parameters, it may look strange for most characters (please see our supplementary material). Nevertheless, as we can still see from Fig.\ \ref{exp:generator} that, the generated face image and the rendered ground truth share a high degree of similarity, even in some regions with complex textures, e.g. the hair. This indicates that our imitator not only fits the training data in a low dimensional face manifold but also learns to decouple the correlations between different facial parameters.\par

\subsection{Facial Similarity Measurement}

\begin{figure}
  \centering
  \includegraphics[width=0.9\linewidth]{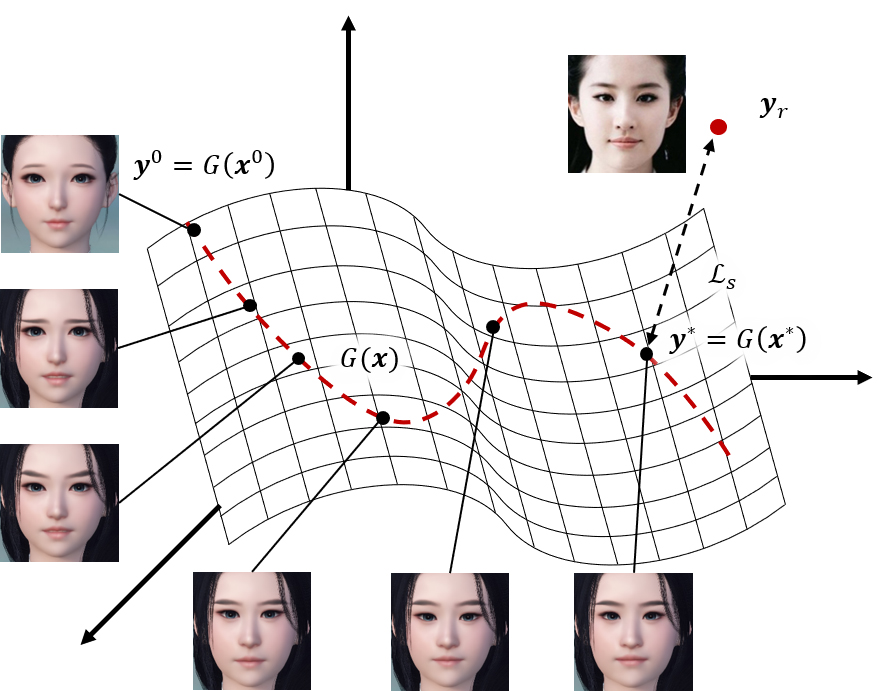}
  \caption{The game character auto-creation can be considered as a searching process on the manifold of the imitator. We aim to find an optimal point ${\bm{y}}^* = G({\bm{x}}^*)$ that minimizes the distance between $\bm{y}$ and the reference face photo $\bm{y}_r$ in their feature space.} 
  \label{manifold}
\end{figure}

Once we have obtained a well-trained imitator $G$, the generation of the facial parameters essentially becomes a face similarity measurement problem. As the input face photo and the rendered game character belong to different image domains, to effectively measure the facial similarity, we design two kinds of loss functions as measurements in terms of both global facial appearance and local details. Instead of directly computing their losses in raw pixel space, we take advantage of the neural style transfer frameworks and compute losses on the feature space that learned by deep neural networks. The parameter generation can be considered as a searching process on the manifold of the imitator, on which we aim to find an optimal point ${\bm{y}}^* = G({\bm{x}}^*)$ that minimizes the distance between $\bm{y}$ and the reference face photo $\bm{y}_r$, as shown in Fig.\ \ref{manifold}. \par

\subsubsection{Discriminative Loss}

We introduce a face recognition model $F_1$ as a measurement of the global appearances of the two faces, e.g. the shape of the face and the overall impression. We follow the idea of the perceptual distance, which has been widely applied in a variety of tasks, e.g. image style transfer \cite{Gatys2016Image}, super-resolution \cite{johnson2016perceptual, ledig2017photo}, and feature visualization \cite{yosinski2015understanding}, and assume that for the different portraits of the same person, their features should have similar representations. To this end, we use a state of the art face recognition model ``Light CNN-29 v2'' \cite{wu2018light} to extract the 256-d facial embeddings of the two facial images and then compute the cosine distance between them as their similarity. The loss function is referred as ``discriminative loss'' since it predicts whether the faces from a real photo and the imitator belong to the same person. The discriminative loss function of the above process is defined as follows:\par
\begin{equation}
\begin{split}
    \mathcal{L}_1(\bm{x}, \bm{y}_r) & = 1 - \cos({F_1(\bm{y}), F_1(\bm{y}_r)}) \\
    & =  1 - \cos({F_1(G(\bm{x})), F_1(\bm{y}_r)}),
\end{split}
\end{equation}
where the cosine distance between two vectors $\bm{a}$ and  $\bm{b}$ is defined as:
\begin{equation}
\cos(\bm{a}, \bm{b})= \frac{<\bm{a}, \bm{b}>}{\sqrt{\|\bm{a}\|_2^2\|\bm{b}\|_2^2}}.
\end{equation}

\subsubsection{Facial Content Loss}

In additional to the discriminative loss, we further define a content loss by computing pixel-wise error based on the facial features extracted from a face semantic segmentation model. The facial content loss can be regarded as a constraint on the shape and displacement of different face components in two images, e.g. the eyes, mouth, and nose. As we care more about modeling facial contents rather than everyday images, the face semantic segmentation network is specifically trained to extract facial image features instead of using off-the-shelf models that are pre-trained on the ImageNet dataset \cite{deng2009imagenet}. We build our facial segmentation model based on the Resnet-50 \cite{he2016deep} where we remove its fully connected layers and increase its output resolution from 1/32 to 1/8. We train this model on the well-known Helen face semantic segmentation dataset \cite{le2012interactive}. To improve the position sensitivity of the facial semantic feature, we further use the segmentation results (class-wise probability maps) as the pixel-wise weights of the feature maps to construct the position-sensitive content loss function. Our facial content loss is defined as follows:
\begin{equation}
    \mathcal{L}_2(\bm{x}, \bm{y}_r) = \|\omega(G(\bm{x}))F_2(G(\bm{x})) - \omega(\bm{y}_r)F_2(\bm{y}_r)\|_1,
\end{equation}
where $F_2$ represents the mapping from input image to the facial semantic features, $\omega$ represents the pixel-wise weights on the features, e.g., $\omega_1$ is the eye-nose-mouth map.\par

\begin{figure}
  \centering
  \includegraphics[width=0.9\linewidth]{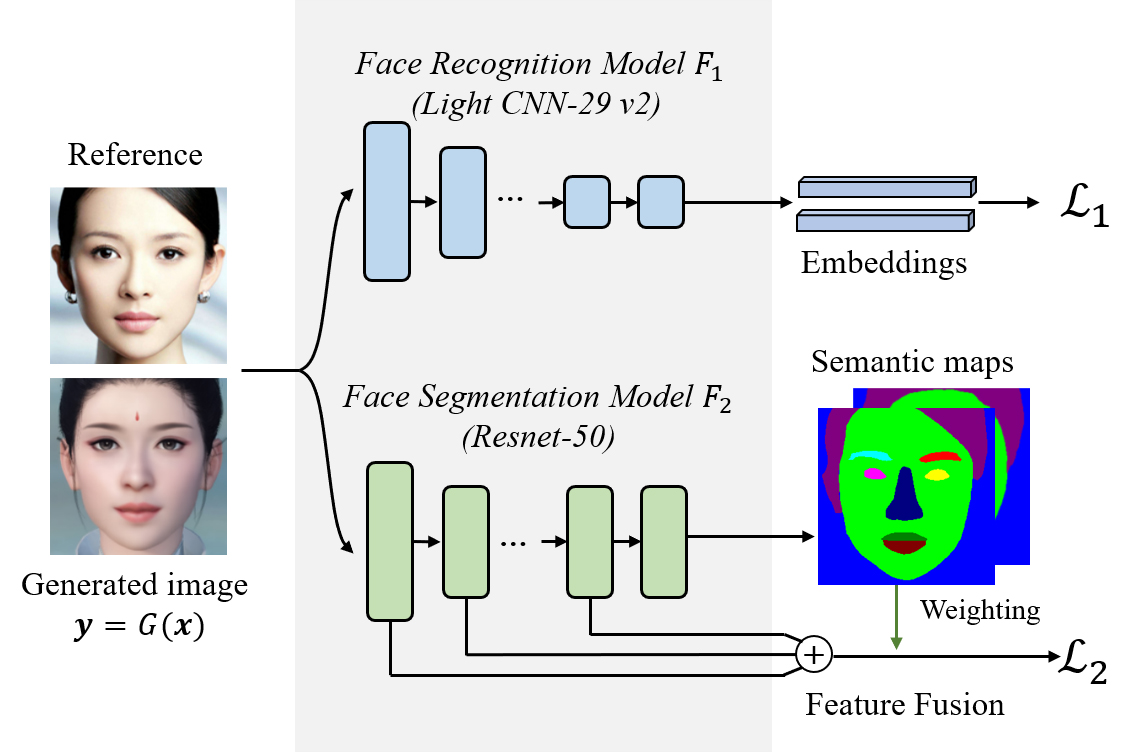}
  \caption{To effectively measure the similarity between two cross-domain face images, we design two kinds of loss functions, i.e.\ the discriminative loss $\mathcal{L}_1$ and the facial content loss $\mathcal{L}_2$, which are defined on the learned feature spaces.
  } \label{extractor}
\end{figure}

The final loss function of our model can be written as a linear combination of the two objectives $\mathcal{L}_1$ and $\mathcal{L}_2$:
\begin{equation}
\begin{split}
\mathcal{L}_{S}(\bm{x}, \bm{y}_r) & = \alpha\mathcal{L}_1 + \mathcal{L}_2 \\
    & =  \alpha (1 - \cos({F_1(G(\bm{x})), F_1(\bm{y}_r)})) \\
    & + \|\omega(G(\bm{x}))F_2(G(\bm{x})) - \omega(\bm{y}_r)F_2(\bm{y}_r)\|_1,
\end{split}
\end{equation}
where the parameter $\alpha$ is used to balance the importance of two tasks. An illustration of our feature extractor is shown in Fig.\ \ref{extractor}. We use the gradient descent method to solve the following optimization problem:
\begin{equation}\label{obj_fac}
\begin{split}
    \min_{\bm{x}} & \ \ \mathcal{L}_{S}(\bm{x}, \bm{y}_r) \\
    \text{s.t.} & \ \ x_i \in [0,1]
\end{split}
\end{equation}
where $\bm{x}=[x_1, x_2, \dots, x_D]$ represents the facial parameters to be optimized and $\bm{y}_r$ represents an input reference face photo. A complete optimization process of our method is summarized as follows:\par

\begin{itemize}
\item \textbf{Stage I.} Train the imitator $G$, the face recognition network $F_1$ and the face segmentation network $F_2$.
\item \textbf{Stage II.} Fix $G$, $F_1$ and $F_2$, initialize and update facial parameters $\bm{x}$, until reach the max-number of iterations: \\
$\bm{x} \leftarrow \bm{x} - \mu \frac{\partial \mathcal{L}_S}{\partial \bm{x}}$ ($\mu$: learning rate). \\
Project $x_i$ to $[0, 1]$: $x_i \leftarrow \max(0, \min(x_i, 1))$.
\end{itemize}\par

\begin{figure*}[ht]
  \centering
  \includegraphics[width=0.9\linewidth]{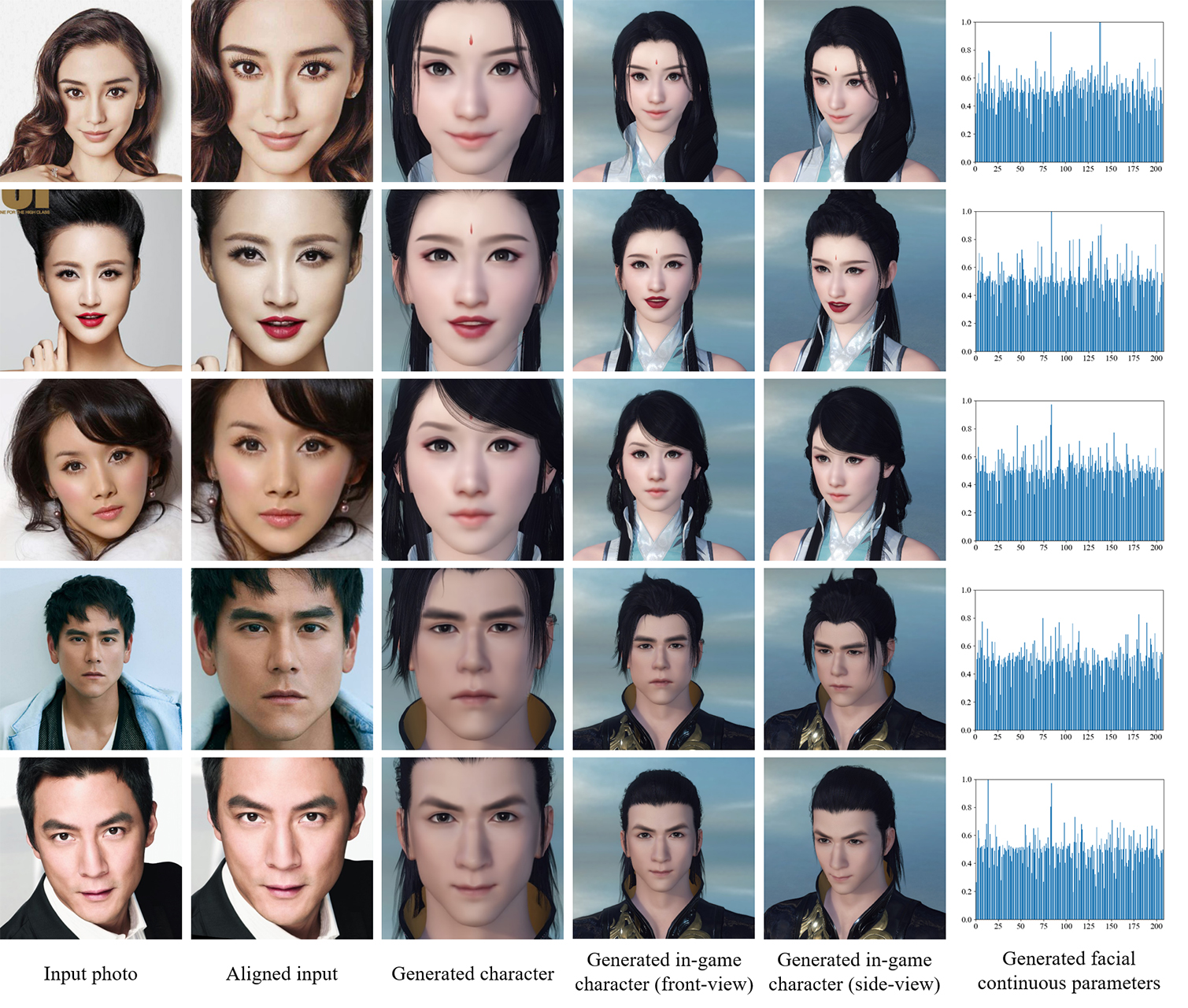}
  \caption{Some input photos and generated characters by using our method (both of the ``identity'' and ``expressions'' are modeled).} 
  \label{exp:big_fig}
\end{figure*}

\subsection{Implementation Details}
\label{implementation}

{\bf Imitator:} In our imitator, the convolution kernel size is set to $4\times 4$, the stride of each transposed convolution layer is set to 2 so that the size of the feature maps will be doubled after each convolution. The Batch-Normalization and ReLU activation are embedded in our imitator after every convolution layers, except for its output layer. Besides, we use the SGD optimizer for training with the $\text{batch\_size}=16$ and $\text{momentum}=0.9$. The learning rate is set to $0.01$, the learning rate decay is set to 10\% per 50 epochs, and the training stops after 500 training epochs.\par

{\bf Facial segmentation network:} We use Resnet-50 \cite{he2016deep} as the backbone of our segmentation network by removing its fully connected layers and adding an additional $1\times 1$ convolution layer at its top. Besides, to increase the output resolution, we change the stride from 2 to 1 at Conv\_3 and Conv\_4. Our model is pre-trained on the ImageNet \cite{deng2009imagenet}, and then fine-tuned on the Helen face semantic segmentation dataset \cite{le2012interactive} with the pixel-wise cross entropy loss. We use the same training configurations as our imitator, except that the learning rate is set to $0.001$.\par

{\bf Facial parameters:} The dimension $D$ of the facial parameters is set to 264 for ``male'' and 310 for ``female''. In these parameters, 208 of them are in continuous values (such as eyebrow length, width, and thickness) and the rest are discrete ones (such as hairstyle, eyebrow style, beard style, and lipstick style). These discrete parameters are encoded as one-hot vectors and are concatenated with the continuous ones. Since the one-hot encodings are difficult to optimize, we use the softmax function to smooth these discrete variables by the following transform:
\begin{equation}
\label{smooth}
h(\bm{x},\beta) = 
\frac{e^{\beta x_k}}{\sum_{i=1}^{D'}e^{\beta x_i}} \ \ k = 1,2,\dots,n,
\end{equation}
where $D'$ represents the dimension of discrete parameters' one-hot encoding. $\beta > 0$ controls the degree of smoothness. We set relatively large $\beta$, say, $\beta=100$, to speed up optimization. We use ``average face'' to initialize the facial parameters $\bm{x}$, i.e.\ we set the all elements in continuous part as 0.5 and set those in discrete part as 0. For a detailed description of our facial parameters, please refer to our supplementary material.\par

{\bf Optimization:} As for the optimization in Stage II, we set $\alpha$ as 0.01, the max-number of iterations as 50, the learning rate $\mu$ as 10 and its decay rate as 20\% per 5 iterations.\par

{\bf Face alignment:} The face alignment is performed (by using dlib library \cite{dlib09}) to align the input photo before it is fed into the feature extractor, and we use the rendered ``average face'' as its reference.\par

\section{Experimental Results and Analysis} 

We construct a celebrity dataset with 50 facial close-up photos to conduct our experiments. Fig.\ \ref{exp:big_fig} shows some input photos and generated facial parameters, from which an in-game character can be rendered by the game engine at multiple views and it shares a high degree of similarity to the input photo. For more generated examples, please refer to our supplementary material. \par

\begin{figure}[ht]
  \centering
  \includegraphics[width=\linewidth]{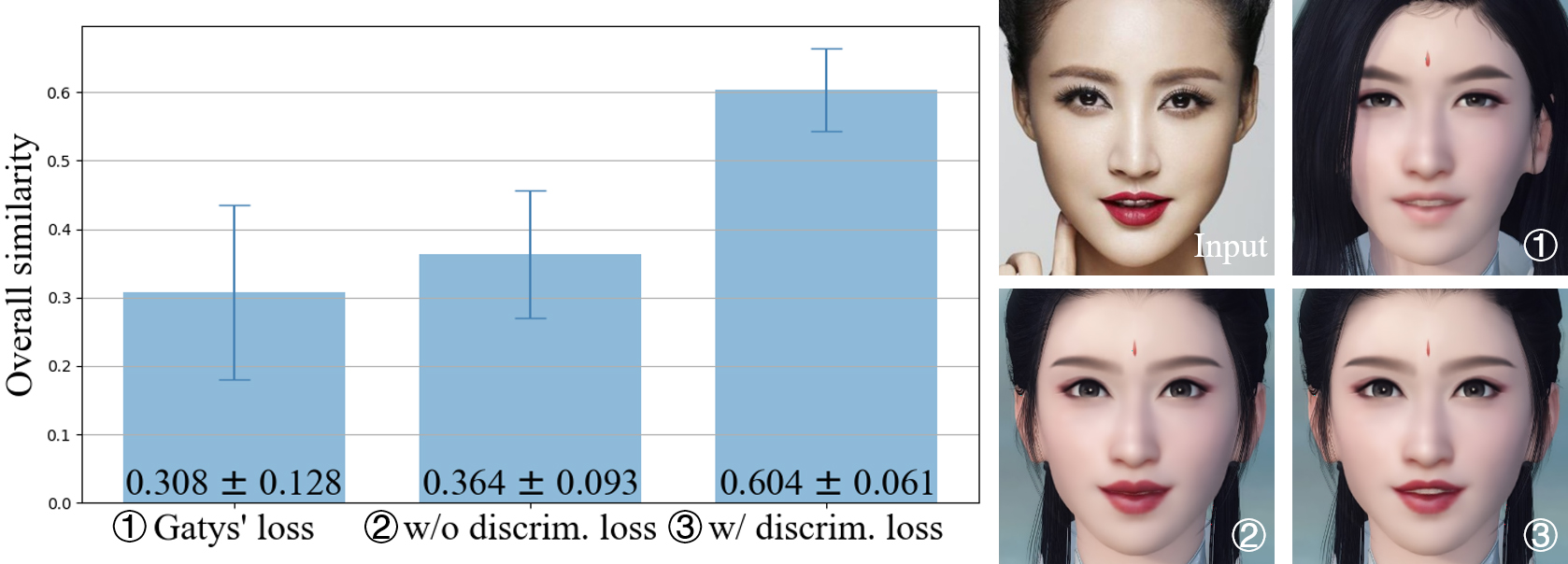}
  \caption{A performance comparison between different objective functions.}
  \label{exp:similarity}
\end{figure}
\subsection{Ablation Studies}

The ablation studies are conducted on our dataset to analyze the importance of each component of the proposed framework, including 1) discriminative loss and 2) facial content loss. \par

{\bf 1) Discriminative loss.} We run our method on our dataset w/ or w/o the help of discriminative loss and further adopt the Gatys' content loss \cite{Gatys2016Image} as the baseline. We compute the similarities between each photo and the corresponding generated results by using the cosine distance on the output of the face recognition model \cite{wu2018light}, as shown in Fig.\ \ref{exp:similarity}. We can observe noticeable similarity improvement when we integrate the discriminative loss.\par

\begin{figure}[ht]
  \centering
  \includegraphics[width=0.8\linewidth]{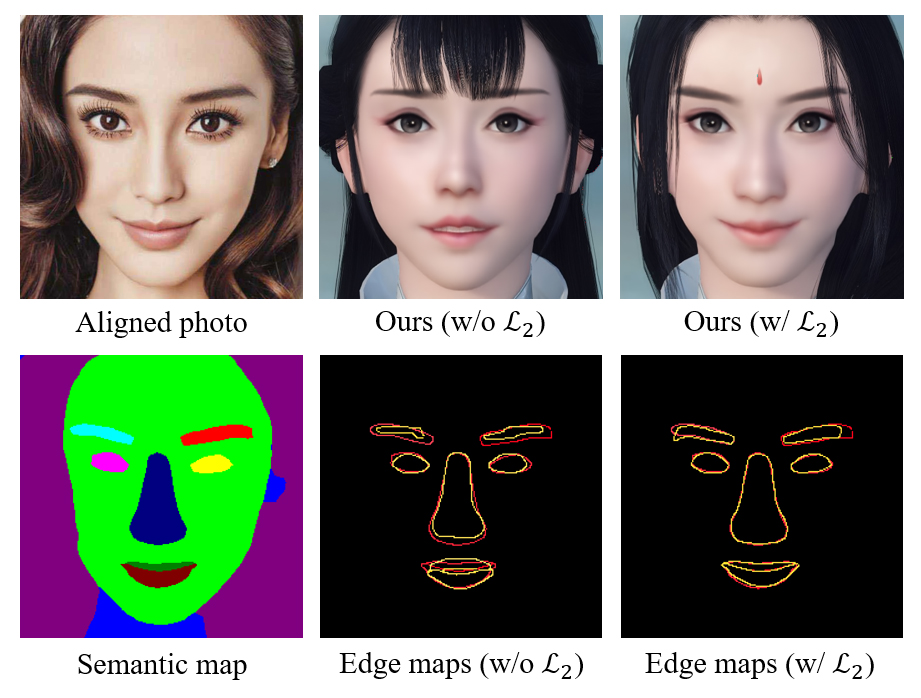}
  \caption{A comparison of the generated faces w/ or w/o the help of the facial content loss. The first column shows the aligned photo and its semantic map produced by our face segmentation model. The second and third columns show the generated faces w/ or w/o the facial content loss. Their edge maps are extracted for a better comparison, where the yellow pixels correspond to the edge of the reference photo and the red pixels correspond to the generated faces.}
  \label{exp:extractor}
\end{figure}

{\bf 2) Facial content loss.} Fig.\ \ref{exp:extractor} shows a comparison of the generated faces w/ or w/o the help the facial content loss. For a clearer view, the facial semantic maps and the edges of facial components are extracted. In Fig.\ \ref{exp:extractor}, the yellow pixels of the edge maps correspond to the edge of the reference photo and the red pixels correspond to the generated faces. We can observe a better correspondence of the pixel location between the input photo and the generated face when we apply the facial content loss.\par

{\bf 3) Subjective evaluation.} To quantitatively analyze the importance of the two losses, we follow the subjective evaluation method used by Wolf \etal \cite{wolf2017unsupervised}. Specifically, we first generate 50 groups of character auto-creation results with different configurations of the similarity measurement loss functions ($\mathcal{L}_1$ only, $\mathcal{L}_2$ only and $\mathcal{L}_1 + \mathcal{L}_2$) on our dataset. We then ask 15 non-professional volunteers to select the best work in each group, of which three characters are in random order. Finally, the selection ratio of an output character is defined as how many percentage it is selected by volunteers as the best one in its group, and the overall selection ratio is used to evaluate the quality of the results. The statistics are shown in Tab.\ \ref{tab:AblationResult}, which indicates that both losses are beneficial for our method. \par

\begin{table}[ht]
\small
\begin{center}
\begin{tabular}{C{2.0cm}C{2.2cm}|C{2.5cm}}
\toprule
\multicolumn{2}{c}{\textbf{Ablations}} & \\
Discrim. $\mathcal{L}_1$ & Facial-Sem. $\mathcal{L}_2$ & Selection Ratio \\
\midrule
$\checkmark$ & $\times$  & $13.47\% \pm 0.38\%$   \\
$\times$ & $\checkmark$ & $36.27\% \pm 0.98\%$  \\
$\checkmark$ & $\checkmark$  &  $ \mathbf{50.26\% \pm 0.40\%}$  \\
\bottomrule
\end{tabular}%
\end{center}
\caption{Subjective evaluation results of two technical components of our method 1) discriminative loss $\mathcal{L}_1$, 2) facial content loss $\mathcal{L}_2$ on our dataset. A higher selection ration indicates better.}
\label{tab:AblationResult}%
\end{table}%

\subsection{Comparison with other methods}

\begin{figure*}
  \centering
  \includegraphics[width=0.9\linewidth]{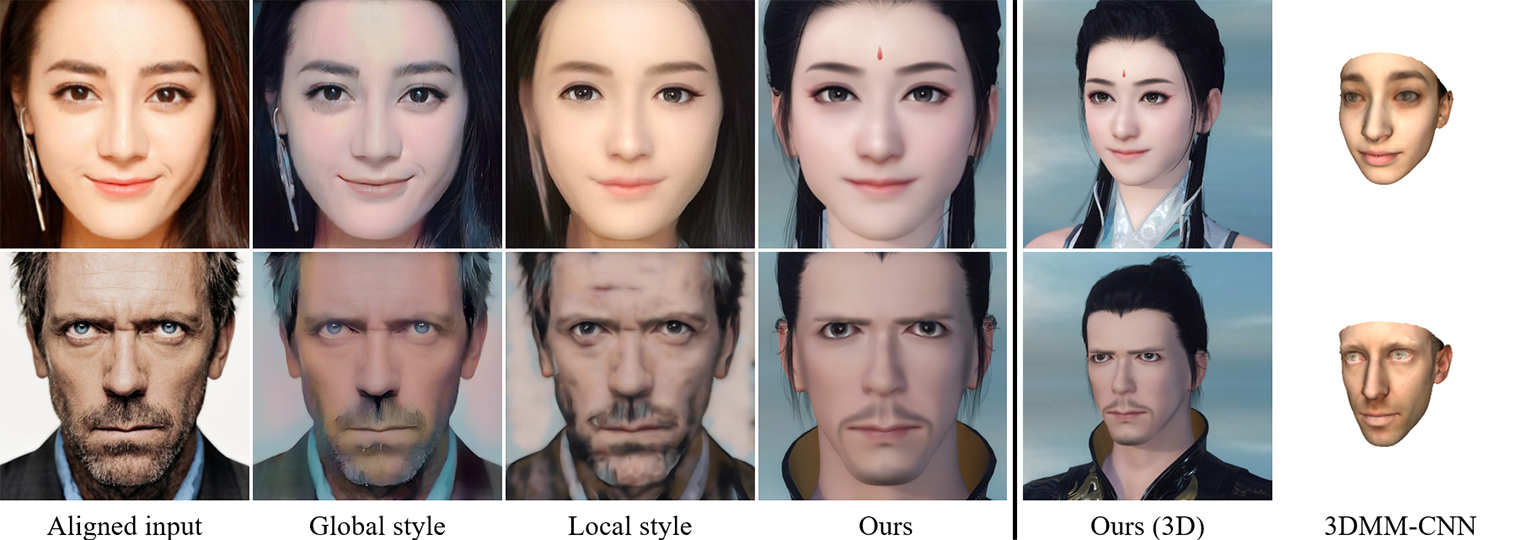}
  \caption{A comparison with other NST methods: Global style \cite{Gatys2016Image} and Local style \cite{Gu_2018_CVPR}, and we use the ``average face'' of each gender as the style reference of these NST methods. We also compare with a popular monocular 3D face reconstruction method: 3DMM-CNN \cite{Tran_2017_CVPR}.}
  \label{exp:compare}
\end{figure*}

\begin{table*}
\centering
\renewcommand\arraystretch{0.8}
\caption{The style similarity and speed performance of different methods. (A higher Mode Score or a lower FID indicates better)}
\begin{tabular}{ccccc}
\toprule
Method & Global style \cite{Gatys2016Image} & Local style \cite{Gu_2018_CVPR} & 3DMM-CNN \cite{Tran_2017_CVPR} & Ours \\
\midrule
Mode Score & $1.0371\pm0.0134$ &$1.0316\pm0.0128$ & -- &$\bm{1.1418\pm0.0049}$\\
\midrule
Fr\'echet Inception Distance & $0.0677\pm0.0018$ & $0.0554\pm0.0025$ & -- &$\bm{0.0390\pm0.0018}$\\
\midrule
Time (run on TITAN Xp) &  22s & 43s  & 15s & 16s\\
\bottomrule
\end{tabular}
\label{indicator}
\end{table*}

We compare our method with some popular neural style transfer methods: global style method \cite{Gatys2016Image} and local style method \cite{Gu_2018_CVPR}. Although these methods are not specifically designed for generating 3D characters, we still compare with them since they are similar with our approach in multiple aspects. Firstly, these methods are all designed to measure the similarity of two images based on deep learning features. Secondly, the iterative optimization algorithms in these methods are all performed at the input of networks. As shown in Fig.\ \ref{exp:compare}, we can see that by separating the image styles from content and reorganizing them, it is difficult to generate vivid game characters. This is because the generated images are not exactly sampled from the game character manifold, thus it's hard to apply these methods in RPG environments. We also compare our method with a popular monocular 3D face reconstruction method: 3DMM-CNN \cite{Tran_2017_CVPR}, as shown in the right side of Fig.\ \ref{exp:compare}. Our auto-created game characters have a high similarity with the inputs while the 3DMM method can only generate masks with similar facial outlines.\par

To quantitatively evaluate the similarity between the generated face and the in-game style reference, we use the Mode Score (MS) \cite{xu2018empirical} and the Fr{\'e}chet Inception Distance (FID) \cite{heusel2017gans} as our metrics. For each test image, we randomly select an image from the imitator training set as its reference and compute the average MS and FID over the entire test set. The above operations are repeated 5 times for computing the final mean value and the standard deviation as listed in Tab.\ \ref{indicator}. The runtime of each method is also recorded. Our method achieves higher style similarity and good speed performance compared with other methods.\par

\subsection{Robustness and limitation.}
\begin{figure}[ht]
  \centering
  \includegraphics[width=0.9\linewidth]{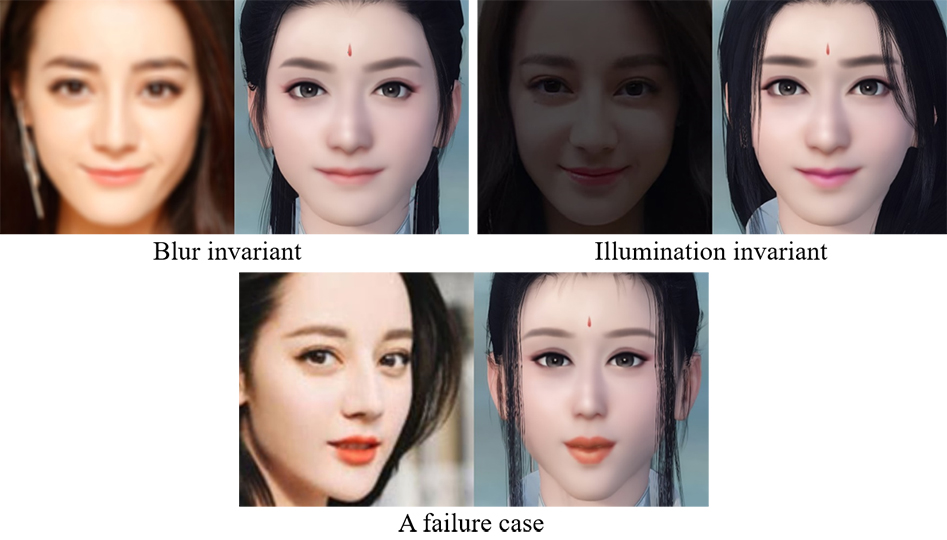}
  \caption{Experiments on robustness.} 
  \label{exp:robustness}
\end{figure}

We further evaluate our method on different blurring and illumination conditions, and our method proves to be robust to these changes, as shown in Fig.\ \ref{exp:robustness}. The last group gives a failure case of our method. Since $\mathcal{L}_2$ is defined on local features, our method is sensitive to pose changes.

\subsection{Generation with artistic portraits}

\begin{figure}[ht]
  \centering
  \includegraphics[width=0.9\linewidth]{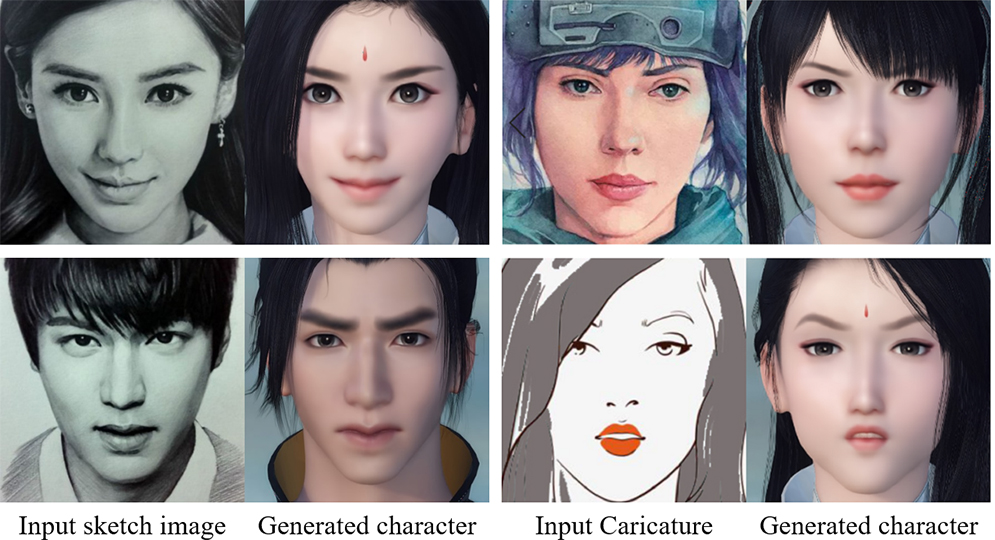}
  \caption{Game character auto-creation on artistic portraits.} 
  \label{exp:cross_domain}
\end{figure}

Not limited to real photos, our method can also generate game characters for some artistic portraits, including the sketch image and caricature. Fig.\ \ref{exp:cross_domain} shows some examples of the generation results. Although the images are collected from a totally different distribution, we still obtain high-quality results since our method measures the similarity based on facial semantics rather than raw pixels.\par

\section{Conclusion}

In this paper, we propose a method for the automatic creation of an in-game character based on an input face photo. We formulate the creation under a facial similarity measurement and a parameter searching paradigm by solving an optimization problem over a large set of physically meaningful facial parameters. Experimental results demonstrate that our method achieves a high degree of generation similarity and robustness between the input face photo and the rendered in-game character in terms of both global appearance and local details.\par


{\small
\bibliographystyle{ieee_fullname}
\bibliography{grimace}
}

\newpage
\appendix
\onecolumn
\section{Appendix}
\subsection{Configurations of our networks}
A detailed configuration of our Imitator $G$ and Face Segmentation Network $F_2$ are listed in Table \ref{table:networks-G} and Table \ref{table:networks-fssn}. As for the details of Face Recognition Network $F_1$, i.e.\ Light CNN-29 v2, please refer to Wu {\etal}'s paper \cite{wu2018light}.\par

Specifically, in a $c\times w \times w / s$ of Convolution / Deconvolution layer, $c$ denotes the number of filters, $w \times w$ denotes the filter's size and $s$ denotes the filter's stride. In a  $w\times w / s$ of Maxpool layer, $w$ denotes the pooling window size, and $s$ denotes the pooling stride. In an $n/s$ Bottleneck block \cite{he2016deep}, $n$ denotes the number of planes, and $s$ denotes the block's stride.\par

\begin{table}[H]
\centering
\renewcommand\arraystretch{1.2}
\begin{tabular}{l|lccc}
    \toprule    
     & \textbf{Layer} & \textbf{Component} & \textbf{Configuration} & \textbf{Output Size}\\
    \midrule
    \multirow{8}{*}{\rotatebox[origin=c]{90}{\textbf{Imitator}}}
     & Conv\_1 & Deconvolution + BN + ReLU & 512x4x4 / 1 & 4x4\\
     & Conv\_2 & Deconvolution + BN + ReLU & 512x4x4 / 2 & 8x8\\
     & Conv\_3 & Deconvolution + BN + ReLU & 512x4x4 / 2 & 16x16\\
     & Conv\_4 & Deconvolution + BN + ReLU & 256x4x4 / 2 & 32x32\\
     & Conv\_5 & Deconvolution + BN + ReLU & 128x4x4 / 2 & 64x64\\
     & Conv\_6 & Deconvolution + BN + ReLU & 64x4x4 / 2 & 128x128\\
     & Conv\_7 & Deconvolution + BN + ReLU & 64x4x4 / 2 & 256x256\\
     & Conv\_8 & Deconvolution & 3x4x4 / 2 & 512x512\\
    \bottomrule
\end{tabular}
    \caption{A detailed configuration of our Imitator $G$.}
    \label{table:networks-G}
\end{table}

\begin{table}[H]
\centering
\renewcommand\arraystretch{1.2}
\begin{tabular}{l|lccc}

    \toprule    
     & \textbf{Layer} & \textbf{Component} & \textbf{Configuration} & \textbf{Output Resolution}\\
    \midrule
    \multirow{7}{*}{\rotatebox[origin=c]{90}{\textbf{Segmentation Model}}}
     & Conv\_1 & Convolution + BN + ReLU   & 64x7x7 / 2 & $\frac{1}{2}\times \frac{1}{2}$\\
     & MaxPool & MaxPool   & 3x3 / 2 & $\frac{1}{4}\times \frac{1}{4}$\\
     & Conv\_2 & 3 x Bottleneck & 64 / 2 & $\frac{1}{8}\times \frac{1}{8}$\\
     & Conv\_3 & 4 x Bottleneck & 128 / 1 & $\frac{1}{8}\times \frac{1}{8}$\\
     & Conv\_4 & 6 x Bottleneck & 256 / 1 & $\frac{1}{8}\times \frac{1}{8}$\\
     & Conv\_5 & 3 x Bottleneck & 512 / 1 & $\frac{1}{8}\times \frac{1}{8}$\\
     & Conv\_6 & Convolution & 11x1x1 / 1 & $\frac{1}{8}\times \frac{1}{8}$\\

    \bottomrule
\end{tabular}
    \caption{A detailed configuration of our Face Segmentation Model $F_2$.}
    \label{table:networks-fssn}
\end{table}

\newpage
\subsection{More examples of generated in-game characters}

\begin{figure}[ht]
    \centering{\includegraphics[width=\linewidth]{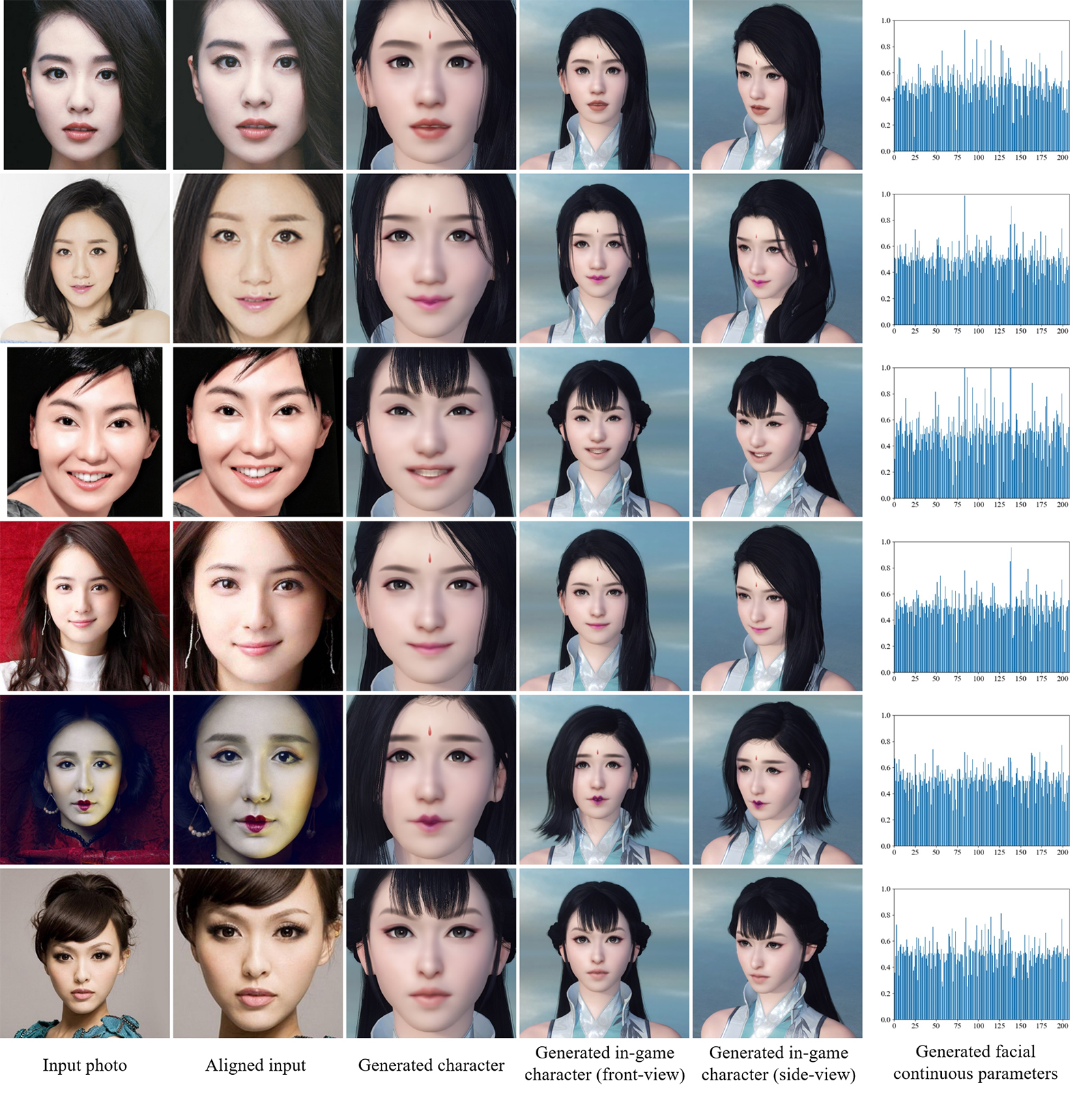}} \\
    \caption{More generated in-game characters (female).}
    \label{fig:female_star}
\end{figure}

\begin{figure}[ht]
    \centering{\includegraphics[width=\linewidth]{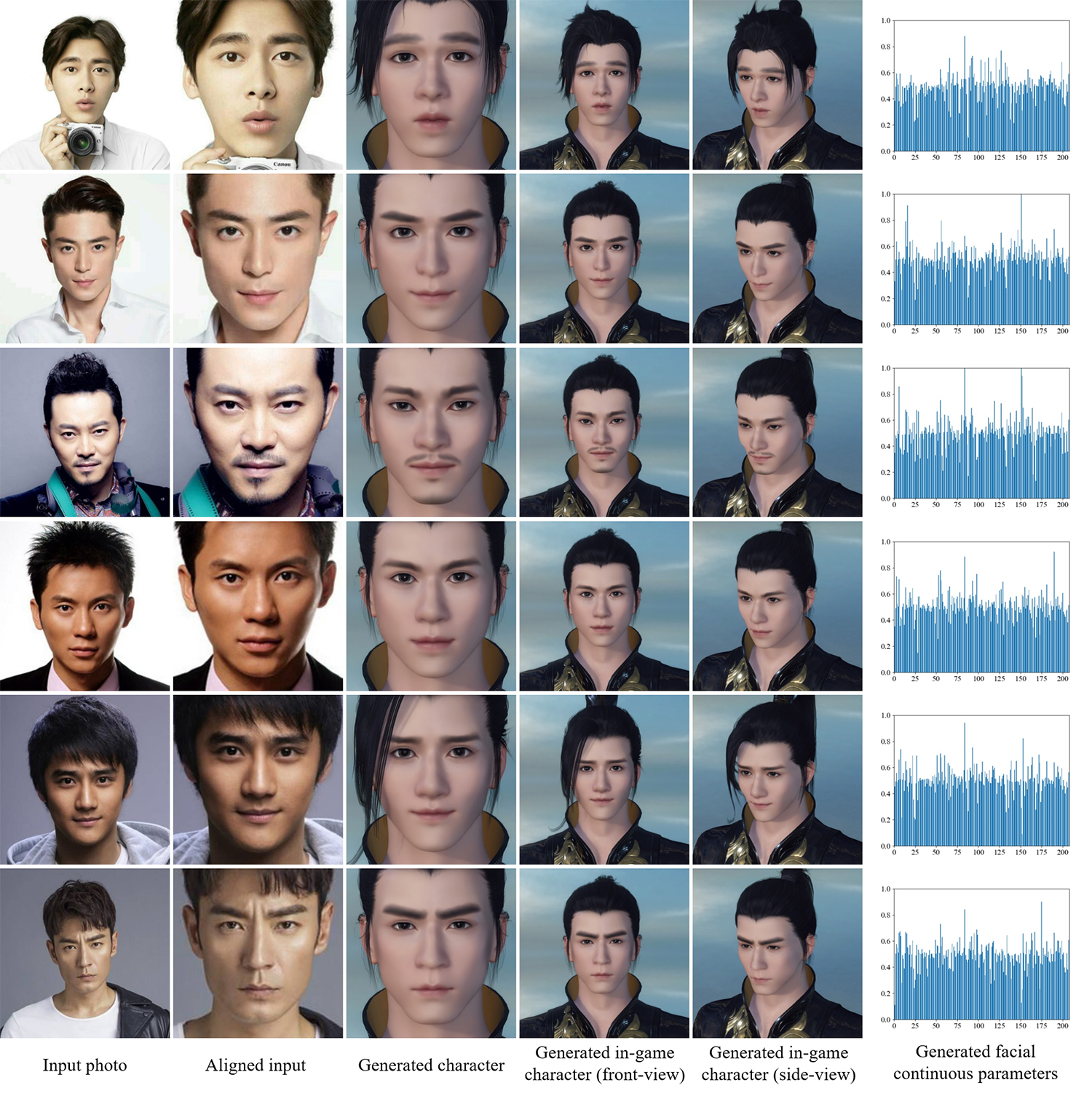}} \\
    \caption{More generated in-game characters (male).}
    \label{fig:male_star}
\end{figure}

\newpage
\subsection{More comparison results}

\begin{figure}[H]
    \centering{\includegraphics[width=\linewidth]{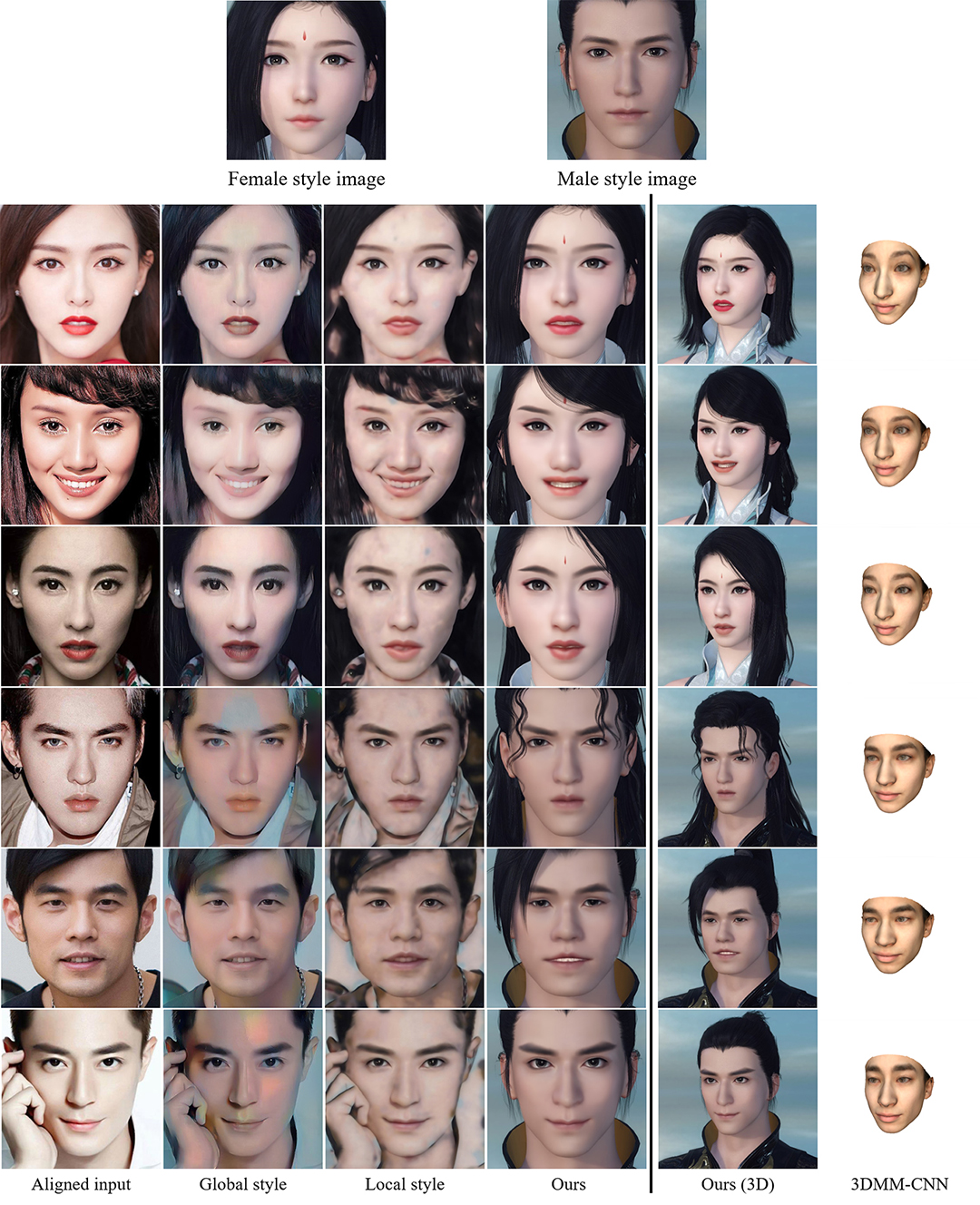}} \\
    \caption{More comparison results with other NST methods: Global style \cite{Gatys2016Image} and Local style \cite{Gu_2018_CVPR}. We use the ``average face'' of each gender as the style images. We also compare with a popular monocular 3D face reconstruction method: 3DMM-CNN \cite{Tran_2017_CVPR}.}
    \label{fig:more_compare}
\end{figure}

\newpage
\subsection{Training samples of our Imitator}
\begin{figure}[H]
    \centering{\includegraphics[width=0.9\linewidth]{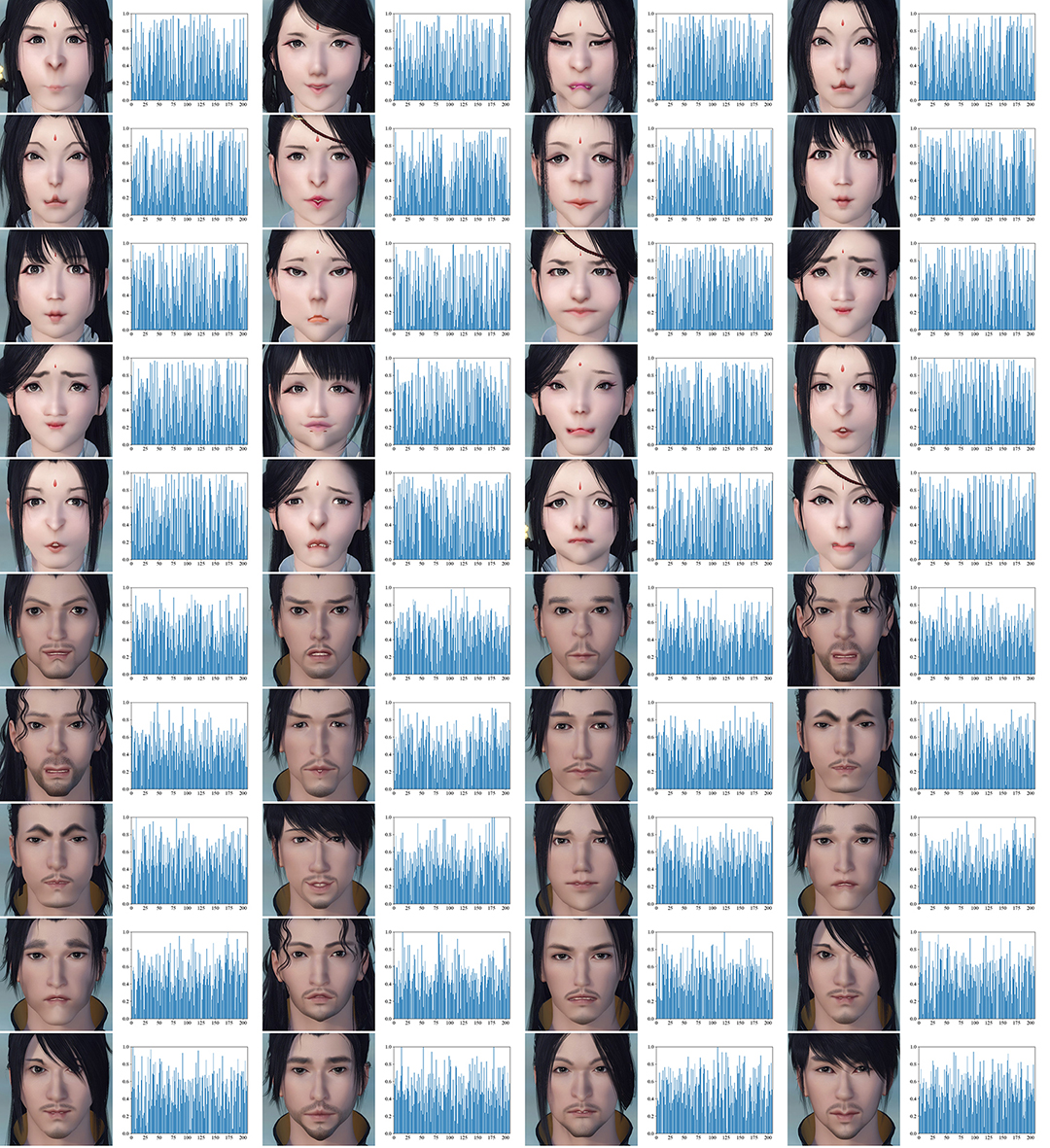}} \\
    \caption{Training samples of imitator and the corresponding facial parameters.}
    \label{fig:training_samples}
\end{figure}

During the training process, we train our imitator with randomly generated game faces other than regular ones, as shown in Fig. \ref{fig:training_samples}. In our experiment, we adopt two imitators to fit female and male 3D models respectively, in order to auto-create characters for different genders.\par

\newpage
\subsection{A description of facial parameters}
Table \ref{table:facial-parameter} lists a detailed description of each facial parameter, where the ``Component'' represents the facial parts which parameters belong to, the ``Controllers'' represents user-adjustable parameters of each facial part (one controller correspond to one continuous parameter), and the ``\# Controllers'' represents the total number of controllers, i.e.\ 208. Besides, there are additional 102 discrete parameters for female (22 hair styles, 36 eyebrow styles, 19 lipstick styles, and 25 lipstick colors) and 56 discrete parameters for male (23 hair styles, 26 eyebrow styles, and 7 beard styles).
\begin{table}[H]
    \centering
    \begin{tabular}{l|l|l|c|r}
        \toprule
        \multicolumn{2}{l|}{Component} & Controllers &  \# controllers & Sum \\
        \midrule
        \multirow{3}{*}{Eyebrow} & eyebrow-head & horizontal-offset, vertical-offset, slope, ... &  8 & \multirow{29}{*}{208}\\
         & eyebrow-body & horizontal-offset, vertical-offset, slope, ... &  8 & \\
         & eyebrow-tail & horizontal-offset, vertical-offset, slope, ... &  8 & \\
         \cline{1-4}
        \multirow{6}{*}{Eye} & whole & horizontal-offset, vertical-offset, slope, ... &  6 & \\
         & outside upper eyelid & horizontal-offset, vertical-offset, slope, ... & 9 & \\    
         & inside upper eyelid & horizontal-offset, vertical-offset, slope, ... & 9 & \\    
         & lower eyelid & horizontal-offset, vertical-offset, slope, ... &  9 & \\   
         & inner corner & horizontal-offset, vertical-offset, slope, ... &  9 & \\   
         & outer corner & horizontal-offset, vertical-offset, slope, ... &  9 & \\
        \cline{1-4}
        \multirow{5}{*}{Nose} & whole & vertical-offset, front-back, slope &  3 & \\
         & bridge & vertical-offset, front-back, slope, ... &  6 & \\    
         & wing & horizontal-offset, vertical-offset, slope, ... &  9 & \\    
         & tip & vertical-offset, front-back, slope, ... &  6 & \\    
         & bottom  & vertical-offset, front-back, slope, ... &  6 & \\   
        \cline{1-4}
        \multirow{6}{*}{Mouth} & whole & vertical-offset, front-back, slope &  3 & \\
         & middle upper lip & vertical-offset, front-back, slope, ... &  6 & \\    
         & outer upper lip & horizontal-offset, vertical-offset, slope, ... &  9 & \\
         & middle lower lip & vertical-offset, front-back, slope, ... &  6 & \\    
         & outer lower lip & horizontal-offset, vertical-offset, slope, ... &  9 & \\
         & corner & horizontal-offset, vertical-offset, slope, ... &  9 & \\     
        \cline{1-4}
        \multirow{9}{*}{Face} & forehead & vertical-offset, front-back, slope, ... &  6 & \\
         & glabellum & vertical-offset, front-back, slope, ... &  6 & \\    
         & cheekbone & horizontal-offset, vertical-offset, slope, ... &  5 & \\    
         & risorius & horizontal-offset, vertical-offset, slope, ... &  5 & \\    
         & cheek & horizontal-offset, vertical-offset, width, ... &  6 & \\     
         & jaw & vertical-offset, front-back, slope, ... &  6 & \\ 
         & lower jaw & horizontal-offset, vertical-offset, slope, ... &  9 & \\ 
         & mandibular corner & horizontal-offset, vertical-offset, slope, ... &  9 & \\ 
         & outer jaw & horizontal-offset, vertical-offset, slope, ... &  9 & \\ 
        \bottomrule
    \end{tabular}
    \caption{A detailed interpretation of each facial parameter (continuous part).}
    \label{table:facial-parameter}
\end{table}

\end{document}